\documentclass[letterpaper, 10 pt, conference]{ieeeconf}  

\IEEEoverridecommandlockouts                              

\usepackage{amsmath}
\usepackage{amsfonts}
\usepackage{amssymb}
\usepackage{mathtools}
\usepackage{graphicx}
\usepackage{color, colortbl}
\usepackage{booktabs}
\usepackage{wrapfig}
\usepackage{listings}
\usepackage{subcaption}
\usepackage{url}
\usepackage{float}

\definecolor{Gray}{gray}{0.9}
\definecolor{LightCyan}{rgb}{0.88,1,1}

\usepackage[hidelinks]{hyperref}
\newcommand{\projecturl}{Project webpage: \href{https://ippon-paper.github.io/}{\textcolor{red}{https://ippon-paper.github.io/}}}

\newcommand\extrafootertext[1]{%
    \bgroup
    \renewcommand\thefootnote{\fnsymbol{footnote}}%
    \renewcommand\thempfootnote{\fnsymbol{mpfootnote}}%
    \footnotetext[0]{#1}%
    \egroup
}

\title{\LARGE \bf
\textbf{IPPON: \textmd{Common Sense Guided} I\textmd{nformative} P\textmd{ath} \linebreak P\textmd{lanning} \textmd{for}  O\textmd{bject Goal} N\textmd{avigation}}}

\author{Kaixian Qu$^{1}$, Jie Tan$^{2}$, Tingnan Zhang$^{2}$, Fei Xia$^{2}$, Cesar Cadena$^{1}$, Marco Hutter$^{1}$
\thanks{This work was supported by an ETH RobotX research grant funded through the ETH Zurich Foundation, by the Swiss National Science Foundation through the National Centre of Competence in Digital Fabrication (NCCR dfab), and by Huawei Tech R\&D (UK) through a research funding agreement. Moreover, this work has been conducted as part of ANYmal Research, a community to advance legged robotics.}
\thanks{$^{1}$ Robotic Systems Lab, ETH Zurich, Switzerland. e-mail: \{kaixqu, cesarc, mahutter\}@ethz.ch. Corresponding author: Kaixian Qu.}%
\thanks{$^{2}$ Google DeepMind, USA. e-mail: \{jietan, tingnan, xiafei\}@google.com.}%
}

\makeatletter
\def\blfootnote{\gdef\@thefnmark{}\@footnotetext}
\makeatother

\begin{document}

\makeatletter
\g@addto@macro\@maketitle{
    \captionsetup{type=figure} 
    \setcounter{figure}{0}      
    \centering                  
    \includegraphics[width=0.9\textwidth]{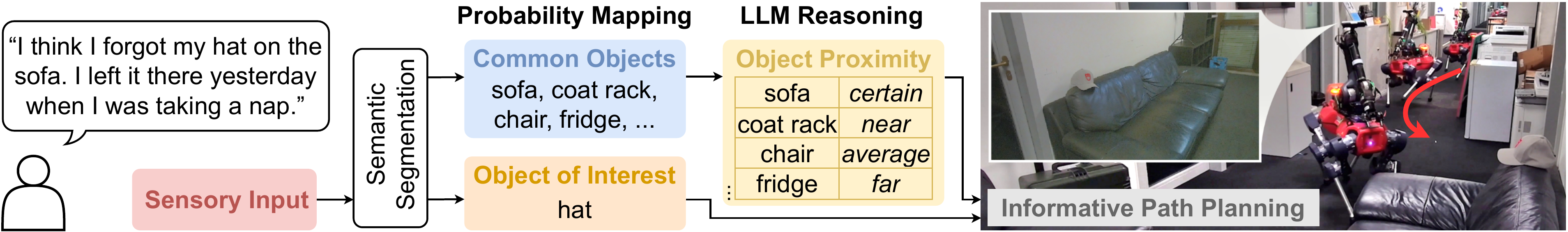} 
    \captionof{figure}{The robot receives a task of finding ``hat'', informed by the context: ``I think I forgot my hat on the sofa. I left it there yesterday when I was taking a nap.'' With the common sense reasoning from an LLM, the robot focuses its search around the sofa area rather than heading towards the kitchen. We visualize the robot's trajectory with a red arrow indicating its direction of movement.} 
    \label{fig:hat-on-sofa}     
    \vspace{-1.0em}               
}
\makeatother

\maketitle


\thispagestyle{empty}
\pagestyle{empty}


\begin{abstract}

Navigating efficiently to an object in an unexplored environment is a critical skill for general-purpose intelligent robots. Recent approaches to this object goal navigation problem have embraced a modular strategy, integrating classical exploration algorithms---notably frontier exploration---with a learned semantic mapping/exploration module. This paper introduces a novel informative path planning and 3D object probability mapping approach. The mapping module computes the probability of the object of interest through semantic segmentation and a Bayes filter. Additionally, it stores probabilities for common objects, which semantically guides the exploration based on common sense priors from a large language model. The planner terminates when the current viewpoint captures enough voxels identified with high confidence as the object of interest. Although our planner follows a zero-shot approach, it achieves state-of-the-art performance as measured by the Success weighted by Path Length (SPL) and Soft SPL in the Habitat ObjectNav Challenge 2023, outperforming other works by more than 20\%. Furthermore, we validate its effectiveness on real robots. \; \projecturl

\end{abstract}


\section{Introduction}
\label{sec:introduction}

Imagine asking an intelligent robot, ``Can you fetch me an apple?'' The robot's response involves initially navigating to find the apple, picking it up, returning to your location, and finally delivering it. In a more practical scenario, the robot lacks prior knowledge of the apple's location and must actively explore a cluttered, unknown environment to locate it. This problem is usually referred to as \emph{Object Goal Navigation}~\cite{batra2020objectnav}, where, given a description of the object of interest (OOI), the agent must decide where to explore and when to terminate within an unknown map. 

The success of a navigation task hinges on three pivotal components: semantic mapping that identifies the placement of objects, planning that determines exploration goals with the shortest traversable paths, and, equally important, semantic guidance that predicts the probable locations of the object of interest with common-sense reasoning. 

The recent development of photo-realistic simulators and reliable object detection/semantic segmentation has matured the semantic mapping technique. This is evident in the substantial growth of the success rate, increasing from 14\% in {Habitat ObjectNav 2020} to 59\% in {Habitat ObjectNav 2023}. This improvement indicates a significant reduction in the likelihood of either failing to detect the OOI or mistakenly identifying other objects as the OOI.

Recent work in object goal navigation has recognized the importance of exploration planners, with frontier-based exploration (FBE)~\cite{yamauchi1997frontier} being the predominant choice. However, the field of autonomous exploration has shifted its focus towards integrating the next best view~\cite{connolly1985determination} with the sampling-based planning (SBP)~\cite{lavalle1998rapidly, karaman2010incremental}. This integration~\cite{bircher2016receding} allows robots to explore 3D spaces, whereas FBE is typically confined to 2D maps.

Semantic guidance is key in helping robots to find objects efficiently. For example, if a robot spots a ``chest of drawers'' on a map, it is more likely to discover a ``bed'' in nearby unexplored areas. Conversely, the presence of a counter suggests that a bed is likely to be somewhere far away. The recent development of large language models (LLMs) has further advanced this area by enabling robots to handle open-vocabulary objects and understand complex contextual instructions, such as ``I think I forgot my hat on the sofa.''

This paper introduces IPPON, which adapts an online informative path planning (IPP) framework~\cite{schmid2020efficient} to address object goal navigation problems. Unlike volumetric exploration, where the gain of each viewpoint is measured by the number of unknown voxels observed, we define gain as the aggregated probability of voxels that are likely part of the OOI. To enable such gain computation, we introduce a 3D object probability mapping algorithm comprising two components: one estimating probabilities for common objects and another for the OOI. The common objects are categories typically present in a given scenario. The planner queries an LLM to obtain common-sense insights on the likely proximity between common objects and the OOI and synthesizes a proximity score for each viewpoint. We illustrate the outline of our approach in Fig.~\ref{fig:hat-on-sofa}.

Our work adheres to a zero-shot approach, i.e., we do not train anything specific on the Habitat ObjectNav dataset~\cite{habitatchallenge2023}. Our work is also open-vocabulary, meaning the user can specify any object of interest, even with descriptive natural language. Leveraging pre-trained foundation models capable of handling both rendered and real images, IPPON demonstrates its effectiveness both in simulation and on real robots. 

The main contributions of this paper are as follows.
\begin{itemize}
    \item We introduce a robust 3D object probability mapping technique using the Bayes filter for object goal navigation. This map representation supports open-vocabulary objects of interest, enables object framing, and is reusable for guiding future object searches.
    \item Leveraging common sense from LLMs, our method generates a proximity map that semantically guides the exploration, notably improving efficiency in common apartment settings.
    \item We extend informative path planning to open-vocabulary object goal navigation and demonstrate that our zero-shot approach outperforms others in the Habitat ObjectNav 2023 Challenge by over 20\%. We also validate its effectiveness on real robots.
\end{itemize}


\section{Related Work}
\label{sec:related-work}

Before the formal introduction of object goal navigation~\cite{batra2020objectnav}, roboticists had long been working on finding objects in known or unknown environments. Early work on unknown environments~\cite{shubina2010visual} proposed greedily selecting viewpoints to maximize the likelihood of observing OOI. Another seminal work~\cite{aydemir2013active} developed a model to balance the exploration of unknown areas and the exploitation of known maps. At that time, object detection was limited, so researchers mostly relied on SIFT-based features~\cite{lowe1999object}, with minimal to no semantic guidance available.

Over the past decade, advancements in computer vision have enabled the finding of diverse objects using RGB-D cameras, and the focus has shifted towards enhancing efficiency through semantic guidance. Researchers have developed models that utilize 2D semantic maps to predict exploration targets with goal-oriented semantic policy~\cite{chaplot2020object}, predict the presence of objects in unseen areas~\cite{georgakis2021learning, ramakrishnan2022poni}, learn graphs of object relationships~\cite{pal2021learning}, or form a correlational object search POMDP, with the correlations between objects provided by domain experts~\cite{zheng2022towards}. Nevertheless, such semantic guidance only operates within a closed set of objects and cannot interpret complex contextual instructions.

The recent approaches have acknowledged the importance of exploration planners, notably using frontier-based exploration (FBE)~\cite{yamauchi1997frontier}. FBE suggests that robots should always navigate to frontiers, defined as the boundary between free and unexplored areas. The current state of the art semantically chooses the best frontier to explore using a learned policy~\cite{chang2023goat, gervet2023navigating} or the commonsense reasoning from LLMs~\cite{shah2023navigation, zhou2023esc} or vision-language models~\cite{yokoyama2023vlfm}. Nonetheless, FBE focuses solely on the gain at the endpoint of each path, overlooking the possibility of gains along the way. In addition, it does not account for the camera's view frustum, which is essential for accurate spatial reasoning.

More recent work in exploration has sought to address these limitations, such as the NBV planner~\cite{bircher2016receding}, which combines sampling-based planning~\cite{lavalle1998rapidly} with the next best view~\cite{connolly1985determination}. NBV assigns a gain to each node based on potential volumetric gain while also considers the total gains accumulated along the path, achieving notable improvements over FBE. Schmid et al.~\cite{schmid2020efficient} extended this approach by framing it as an online informative path planning (IPP) method and demonstrating its effectiveness in 3D reconstruction. However, this online IPP has yet to be applied to tasks that require termination and rely heavily on semantic information, such as object goal navigation.

\section{Problem Formulation}
\label{sec:formulation}

We adhere to the Habitat ObjectNav Challenge 2023 problem setup~\cite{habitatchallenge2023}, where the agent takes actions in a continuous velocity space, including moving forward, turning left or right, and looking up or down (i.e., tilting the camera). After each action, the agent receives an observation that includes an RGB-D image and its current state, and then decides on the next action. An episode is considered successful if the agent triggers a termination signal when an OOI is within 1 meter and can be seen by turning around and tilting the camera. Failing to trigger the termination signal correctly or within the step or time limit is deemed a failure.


\section{Method}
\label{sec:method}

\begin{figure*}[t]
    \centering
    \includegraphics[width=0.90\textwidth]{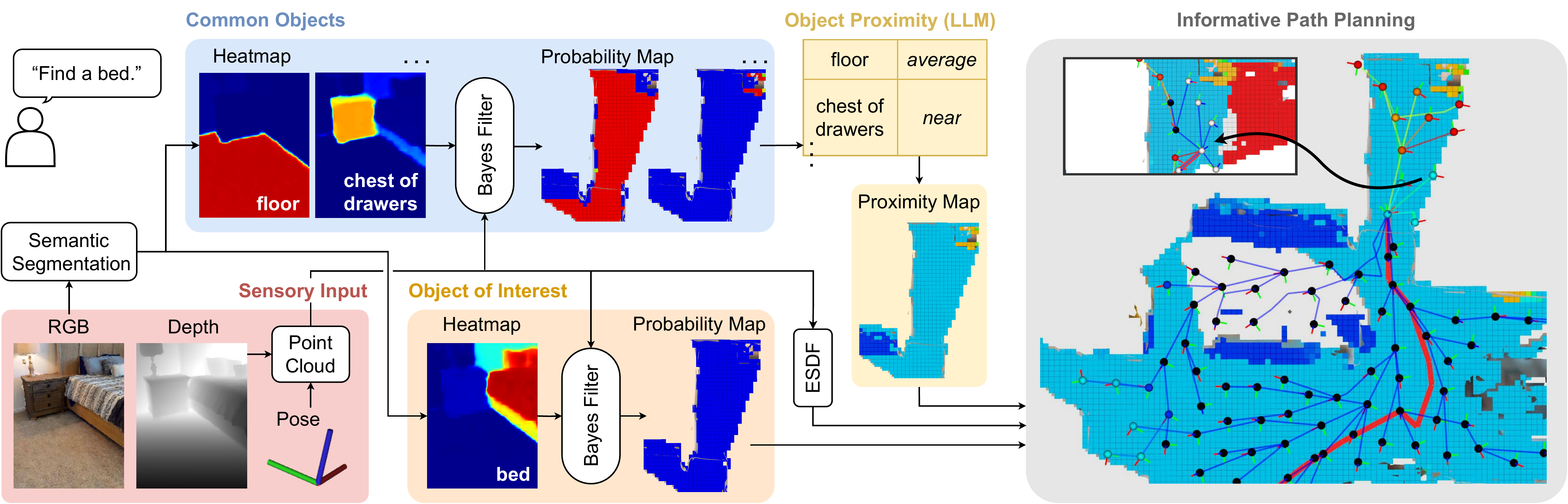}
    \caption{The pipeline of IPPON consists of three main components: 3D object probability mapping using the Bayes Filter, common sense reasoning that provides a proximity map for the OOI, and informative path planning based on the probability map combined with the proximity map. IPPON uses Voxblox~\cite{oleynikova2017voxblox} to compute the Euclidean signed distance field (ESDF) online for traversability estimation. Notice that we are showing the map and plan from the previous step, so the detected bed is not yet reflected on the probability map. We color-code the nodes in IPP: terminating nodes in white, minimally-exploring nodes in black, and the rest based on their gain -- from highest (red) to lowest (blue). In this example of finding a bed, the robot initially detects a chest of drawers in the top right corner. Due to its close proximity, the planner subsequently focuses on exploring the nearby area (red nodes), leading it to terminate in front of the bed. The final map and plan are shown in the black-outlined image. }
    \label{fig:ippon-pipeline}
    \vspace{-1.5em}
\end{figure*}

We illustrate the IPPON pipeline in Fig.~\ref{fig:ippon-pipeline}, where IPPON begins by performing semantic segmentation with the SAN model~\cite{xu2023side}. SAN is an open-vocabulary semantic segmentation model that takes a list of labels and an RGB image as inputs, and outputs the probability of each pixel belonging to these labels (represented as heatmaps). Using an open-vocabulary model requires an adequate number of labels, which we refer to as common objects, to ensure that most elements in the image are correctly identified. For the Habitat ObjectNav Challenge, we naturally use the provided 40 semantic labels; in real-world experiments, we manually select significant objects. These pixel probabilities are then merged into 3D voxels using the Bayes filter to estimate the posterior probabilities of each voxel associated with different objects. We query an LLM to evaluate the proximity between common objects and the OOI and create a proximity score map that semantically guides the informative path planning to decide exploration targets. Further details on these modules are provided below.

\subsection{Object Probability Mapping}

The mapping algorithm dynamically updates the probability that voxel $v$ belongs to a common object category $\mathcal{C}_j$ or the object of interest $\mathcal{O}$, considering all image observations $\mathbf{I}_{1:k}$ up to time $k$. We illustrate this computation with the Bayes filter for $\mathcal{O}$, but the same principle applies to $\mathcal{C}_j$.
\begin{equation}
  p(v \in \mathcal{O} \mid \mathbf{I}_{1:k}) = \frac{p(\mathbf{I}_{k} \mid v \in \mathcal{O}) \, p(v \in \mathcal{O} \mid \mathbf{I}_{1:k-1})}{p(\mathbf{I}_{k} \mid \mathbf{I}_{1:k-1})}.
  \label{eq:bayes-rule}
\end{equation}
However, the observation model $p(\mathbf{I}_{k} \mid v \in \mathcal{O})$ is unavailable, as the semantic segmentation network provides $p(v \in \mathcal{O} \mid \mathbf{I}_{k})$. The relationship between these probabilities can be obtained using another application of Bayes' rule, which leads to the following expression:
\begin{equation}
  p(v \in \mathcal{O} \mid \mathbf{I}_{1:k}) = \frac{1}{Z} \frac{p(v \in \mathcal{O}\mid \mathbf{I}_{k}) \, p(v \in \mathcal{O} \mid \mathbf{I}_{1:k-1})}{p(v \in \mathcal{O})},
\end{equation}
where $Z = p(\mathbf{I}_{k} \mid \mathbf{I}_{1:k-1}) / p(\mathbf{I}_{k})$ is the normalization constant. Following similar approaches as in ~\cite{dewan2020deeptemporalseg}, we assume the prior probabilities are equal across all categories, i.e., $p(v \in \mathcal{O}) = p(v \in \mathcal{C}_0) = p(v \in \mathcal{C}_1) = \ldots = p(v \in \mathcal{C}_N)$. In subsequent sections, with a slight abuse of notation, we use $p(v \in \mathcal{O})$ to denote the posterior probability (not the prior), omitting conditioning on observations for simplicity.

Bayes filters play a key role in building accurate semantic maps by correcting misclassifications over multiple iterations of Bayesian updates. This helps mitigate issues like flickering or inconsistency in semantic segmentation. Furthermore, Bayes filters iteratively refine object probabilities, generally increasing the likelihood of detected objects towards 1.0, while reducing that of undetected objects towards 0.0.

One key difference between mapping OOI and common objects lies in the distance range that we fuse the pixel probability into 3D. The maximum distance is generally smaller for OOI because they typically require closer inspection for accurate semantic segmentation, as incorrect predictions may cause the robot to stop at the wrong object or mistakenly assume its absence and thus avoid approaching it further. However, in the Habitat ObjectNav Challenge, where the depth camera features a narrow field of view and limited depth range, we directly set the maximum mapping distance for all objects to the maximum depth ($5$ m).

\subsection{Semantic Guidance}
\label{subsec:common-sense}

The goal of semantic guidance in our framework is to compute an imagined probability $p_{img}$ for each node, which represents the likelihood that an unknown voxel belongs to the OOI within its view frustum. Similar to~\cite{shah2023navigation, zhou2023esc}, we query an LLM (GPT-4~\cite{achiam2023gpt}) to evaluate the proximity of each common object to the OOI, categorizing it into four levels: \textit{certain}, \textit{near}, \textit{average}, \textit{far}. The rough outline for the prompt is listed in Listing~\ref{lst:llm_prompt}, and the full template and the example responses can be found on the \href{https://ippon-paper.github.io/}{project webpage}. Note that LLMs have the capability to understand additional contextual information, as illustrated in Fig.~\ref{fig:hat-on-sofa}.

\vspace{-0.5em}
\begin{lstlisting}[captionpos=t, caption={Prompt for querying object proximity levels.}, label={lst:llm_prompt}]
I am an assistant autonomous robot trying to help my owner find a {ooi} (object). {context_information}.

My search is within {scenario}, which has the following list of common categories: {common_objects}.

Now I need your help to search for objects more efficiently. For each category, please assign its proximity level to a {ooi}, indicating how likely it is to find a {ooi} nearby if I am close to that category.
\end{lstlisting}

\begin{figure}[t]
    \centering
    \includegraphics[width=\linewidth]{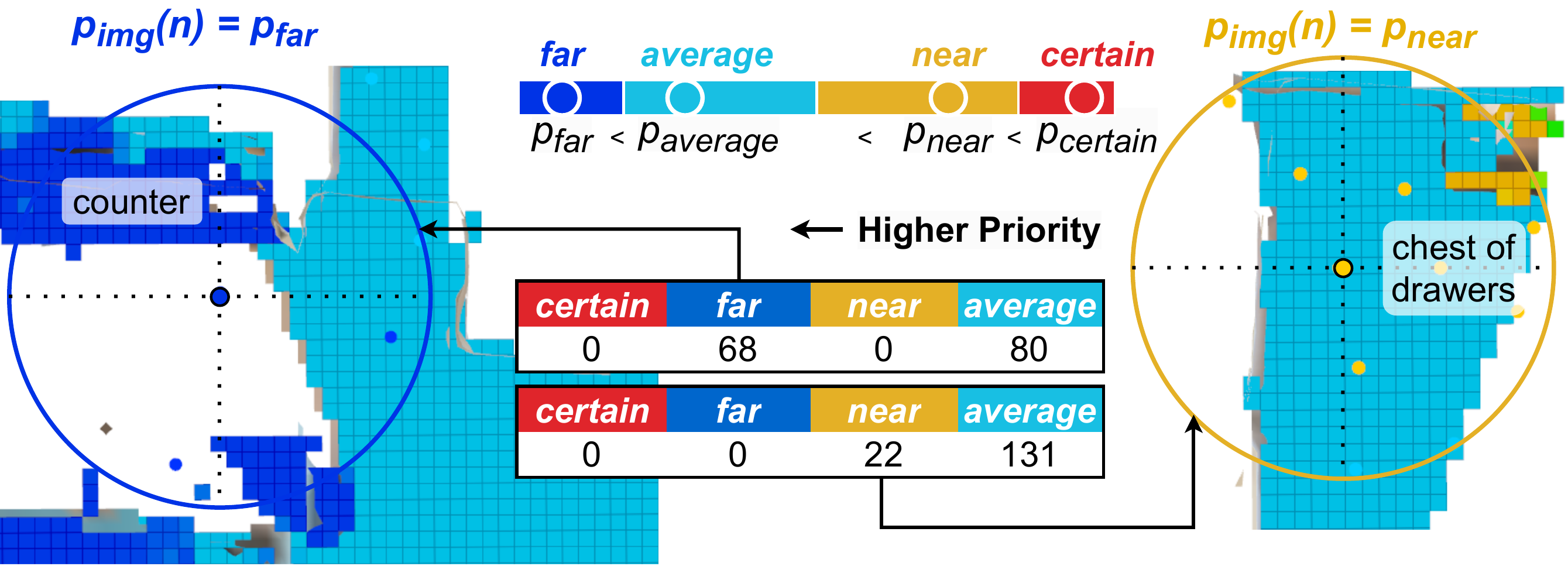}
    \caption{Calculation of $p_{img}$ for the scene in Fig.~\ref{fig:ippon-pipeline}: the orange node near a chest of drawers has a high probability ($p_{near}$) of locating a bed, while the blue node near the counter has a low probability ($p_{far}$).}
    \label{fig:imagined-probability}
    \vspace{-2em}
\end{figure}

We then transform the proximity level into a probability $l(\mathcal{O} \mid \mathcal{C}_j)$ based on predefined probability levels where $p_{certain} > p_{near} > p_{average} > p_{far}$. These probabilities then form a weighted average at the voxel level, indicating the likelihood of locating the OOI $\mathcal{O}$ near that voxel:
\begin{equation}
l(v) = \sum\nolimits_{j = 1}^{N} l(\mathcal{O} \mid \mathcal{C}_j) \, p(v \in \mathcal{C}_j).
\end{equation}
Finally, each viewpoint counts the nearby visible voxels linked to each proximity level and selects the highest-ranking non-empty category, following the prioritization order of \textit{certain}, \textit{far}, \textit{near}, \textit{average}, to determine the imagined probability $p_{img}$ (see Fig.~\ref{fig:imagined-probability} for an example). Figure~\ref{fig:ippon-pipeline} illustrates how this common sense assists in finding the OOI.

The observed probability of common objects can be effectively utilized for subsequent object searches. By querying the LLM for the proximity between the new OOI and existing common objects, a new proximity map is rapidly generated to guide the search. This map can remain useful over a long horizon when common objects are typically stationary, like sofas or tables.

\subsection{Informative Path Planning}
\label{subsec:ipp}

The informative path planning~\cite{schmid2020efficient} grows a tree with sampling-based planning. Each node, representing a viewpoint in 3D space, is associated with a gain, while each edge, representing a traversable path between nodes, is assigned a cost of traveling time. The planning algorithm can select the next optimal target based on various criteria, such as the gain-to-cost ratio as proposed in~\cite{schmid2020efficient}. Upon reaching the next node along the optimal path, the planner sets this node as the new root and rewires the tree. Then, it updates the gain and cost and re-evaluates the next optimal target.

\subsubsection{Traversability Estimation}
\label{subsubsec:collision-checking}

A pose or a path is considered untraversable if the robot's body collides with the environment or the base fails to maintain contact with the ground. The robot is approximated by a list of collision spheres, and the distance is checked using the Euclidean signed distance field $sdf(\cdot)$ from Voxblox~\cite{oleynikova2017voxblox}. Since the robot does not have prior knowledge about the map, it initially looks down to estimate its height (ranging between 1.30 and 1.42 meters in the Habitat ObjectNav Challenge) and then looks around to understand its vicinity. The robot maximizes its camera pitch movement (looking up/down) when traveling to perceive as much as possible. However, the robot may still collide with the environment due to mesh artifacts; when this happens, the planner marks the triangular region ahead as untraversable and finds an alternative path.
\subsubsection{Node Evaluation}
\label{subsubsec:node-evaluation}

For each pose $n$, we conduct ray tracing on the 3D map to detect all visible voxels, denoted by $\mathcal{V}(n)$. The gain for each node is determined by the total probability of the voxels in $\mathcal{V}(n)$ are the OOI, considering its occupancy state,
\begin{equation}
    G(n) = \sum_{v \in \mathcal{V}(n)} \big( \mathbf{1}_{v \in \mathcal{V}_{\mathcal{O}}} \cdot p(v \in \mathcal{O}) + \mathbf{1}_{v \notin \mathcal{V}_{\mathcal{O}}} \cdot p_{img}(n) \big) \cdot p_{occ}(v),
    \label{eq:gain-computation}
\end{equation}
\normalsize
where $\mathbf{1}$ is the indicator function and $\mathcal{V}_{\mathcal{O}}$ contains voxels mapped with the OOI probability $p(v \in \mathcal{O})$. $p_{img}$ is the imagined probability from the semantic guidance (see Sec.~\ref{subsec:common-sense}). The probability of a voxel being occupied, $p_{occ}$, is mapped to 0 if it is known to be free, 1 if occupied, and to a user-defined occupied probability if its state is unknown.

In addition to exploration, the robot must terminate in front of the OOI. Hence, we implement an additional termination evaluation that counts the number of nearby visible voxels whose OOI probability is higher than a non-class-specific threshold $p_T$ (referred to as OOI voxels),
\begin{equation}
    T(n) = \sum_{v \in \mathcal{N}_{T}(n)} \mathbf{1}_{v \in \mathcal{V}_{\mathcal{O}}} \cdot \mathbf{1}_{p(v \in \mathcal{O}) > p_{T}} \cdot p_{occ}(v).
\end{equation}
The terminating nodes $\mathcal{T}$, where the robot can conclude navigation, is defined as the set of all nodes whose $T$ value is higher than $T_{min}$. $T_{min}$ varies depending on the physical size of the OOI (smaller for plants and larger for beds). $\mathcal{N}_{T}(n)$ represents all the voxels visible within a 1-meter radius from $n$, which the robot can observe by turning around or looking up/down. This aligns with the success criteria in the Habitat ObjectNav Challenge. 

Following the methodologies in \cite{schmid2020efficient,witting2018history}, we sample position coordinates $(x, y)$ of nodes and optimize their yaw and pitch for better gain and termination. The position coordinates are first sampled densely within a local region before expanding to the remaining space. To enable object framing, we set the orientation for terminating nodes based on the highest number of OOI voxels in $\mathcal{N}_{T}(n)$. For nodes that do not explore much, we set their orientation to the direction of travel to mitigate the turning costs.


\subsubsection{Node Connection}
\label{subsubsec:node-connecction}

\begin{figure}[t]
    \centering
    \includegraphics[width=0.95\linewidth]{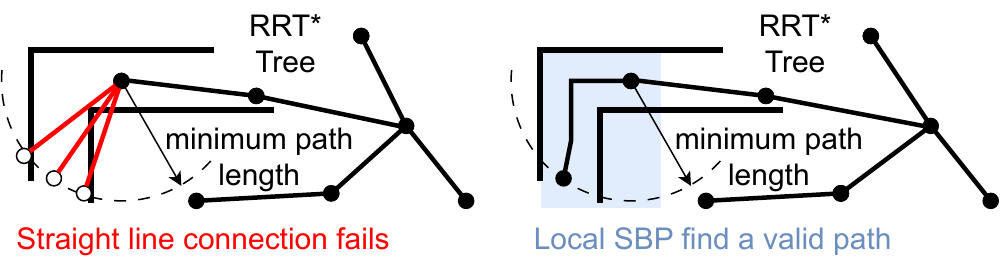}
    \caption{The local SBP can find a path in a narrow ``L"-shaped corridor when a straight-line connection fails.}
    \label{fig:local-sbp}
    \vspace{-2em}
\end{figure}

Once a new node has been evaluated, the planner adds it to the existing tree using the RRT* algorithm~\cite{karaman2010incremental}. Specifically, the planner first connects it to a neighboring node and then rewires neighboring nodes to create a tree with asymptotically minimum cost to every node. The connection verification typically only validates a straight-line path; if this check fails, the path is deemed invalid. However, in the informative path planning, nodes or viewpoints are deliberately spaced apart because it assumes viewpoints have independent gains from each other~\cite{schmid2020efficient}. Such spacing can render direct connections infeasible, especially in constrained environments like narrow ``L"-shaped corridors. Therefore, we introduce another local sampling-based planner (SBP) from OMPL library~\cite{sucan2012open} to explore local alternative routes between two viewpoints when the straight-line connection fails (see Fig.~\ref{fig:local-sbp}). This local SBP minimizes the sum of the path length, penalized by the inverse of the path clearance (from obstacles).

\subsubsection{Node Selection}
\label{subsubsec:node-selection}

When selecting the target node, the planner aims to maximize efficiency, particularly the gain-to-cost ratio, and gives priority to terminating nodes $\mathcal{T}$,
\begin{equation}
n^* =
  \begin{cases} 
    \arg \max\nolimits_{n \in \mathcal{T}} \frac{T(n)}{C(P_n)} & \text{if } \mathcal{T} \neq \emptyset \\
    \arg \max\nolimits_{n} \hspace{12px} \frac{G(P_n)}{C(P_n)} & \text{otherwise}
  \end{cases}.
  \label{eq:node-selection}
\end{equation}
Here, $P_n$ denotes the path from the root to node $n$, with $G(P_n)$ and $C(P_n)$ representing the total gain and cost along path $P_n$, respectively.


\section{Results}
\label{sec:result}

\subsection{Habitat ObjectNav Evaluation}

We evaluated our method on the Habitat ObjectNav 2023 Challenge, standard phase (1000 episodes)~\cite{habitatchallenge2023}. Compared to the previous year's challenge, the action space of this year has transitioned from discrete to continuous, aligning well with the assumptions of sampling-based planning. The considered categories in this challenge are fixed: ``chair'', ``bed'', ``plant'', ``toilet'', ``tv'', and ``sofa''. We show our evaluation result in Table~\ref{tab:object2023-evaluation}.

\begin{table}[h]
  \vspace{-0.5em}
  \centering
    \centering
  \caption{Evaluation results on the Habitat ObjectNav 2023 Challenge (standard phase).}
    \begin{tabular}{l*{3}{c}}
    \toprule
    Method & SPL  $\uparrow$ & Soft SPL  $\uparrow$ & Success $\uparrow$ \\
    \midrule
    Host team & 0.05 & 0.27 & 0.12 \\
    \rowcolor{Gray}
    Auxiliary RL~\cite{ye2021auxiliary} & 0.10 & 0.31 & 0.18 \\
    ICanFly & 0.26 & 0.37 & 0.43  \\
    \rowcolor{Gray}
    SkillFusion~\cite{staroverov2023skill} & 0.28 & 0.34 & 0.53 \\
    SkillTron~\cite{zemskova2024interactive} & 0.28 & 0.36 &  \textbf{0.59} \\
    \midrule
    IPPON (ours) & \textbf{0.34} & \textbf{0.46} & 0.54 \\
    \bottomrule
    \end{tabular}
  \label{tab:object2023-evaluation}
  \vspace{-1em}
\end{table}

Unlike other teams that are all trained on this dataset\footnote{Except ICanFly as there is no other information. The result of Auxiliary RL is taken from~\cite{zemskova2024interactive}, and the results of SkillFusion and ICanFly are taken from the leaderboard~\cite{habitatchallenge2023}.}, our method follows a zero-shot approach. Still, it achieves the best performance in the Success weighted by Path Length (SPL) and Soft SPL, outperforming the previous state of the art by more than 20\%.  In addition, our planner supports open-vocabulary OOI, whereas SkillTron~\cite{zemskova2024interactive}, SkillFusion~\cite{zemskova2024interactive}, and Auxiliary RL~\cite{ye2021auxiliary} all use a dedicated fixed-class semantic segmentation network.

\subsection{Hardware Experiments}

For the real-world experiments, we use a legged robot ANYmal~\cite{hutter2016anymal} with an arm mounted on top. A ZED X camera, featuring a 74$^{\circ}$ horizontal FoV, is mounted on the elbow of the arm. The robot's pose is estimated by a SLAM system~\cite{nubert2022graph}. All computations are performed onboard the robot, except for querying the LLM for proximity levels. The Jetson Orin handles heatmap computation and ZED depth rendering, while mapping and planning modules run on an Intel Core i7 CPU. The videos of our experiments can be found on the \href{https://ippon-paper.github.io/}{project webpage}.


We conducted tests on ten objects, with three trials each, covering four basic categories from the Habitat Challenge (``plant,'' ``chair,'' ``tv,'' ``sofa''), four open-vocabulary categories (``toy elephant,'' ``screwdriver,'' ``microwave oven,'' ``coffee machine''), and two context-specific categories (``hat'' associated with ``left it on the sofa,'' ``student card'' linked to ``I just use it on the printer''). We achieved 26 successes out of these 30 tests, with a success rate of 0.87. In cases where the robot is given the context ``I left my hat on the sofa,'' it quickly explores the area near the sofa, as illustrated in Fig. 1. For the failures, there was one instance when the robot terminated in front of a yellow button when the OOI was ``screwdriver.'' This occurred because the button did not align well with any of the predefined common objects, with the closest match being the OOI (screwdriver). The other three failures happened because the object (student card) was too small to be recognized by the semantic segmentation network unless positioned very close. 

\begin{figure}[t]
    \centering
    \includegraphics[width=0.95\columnwidth]{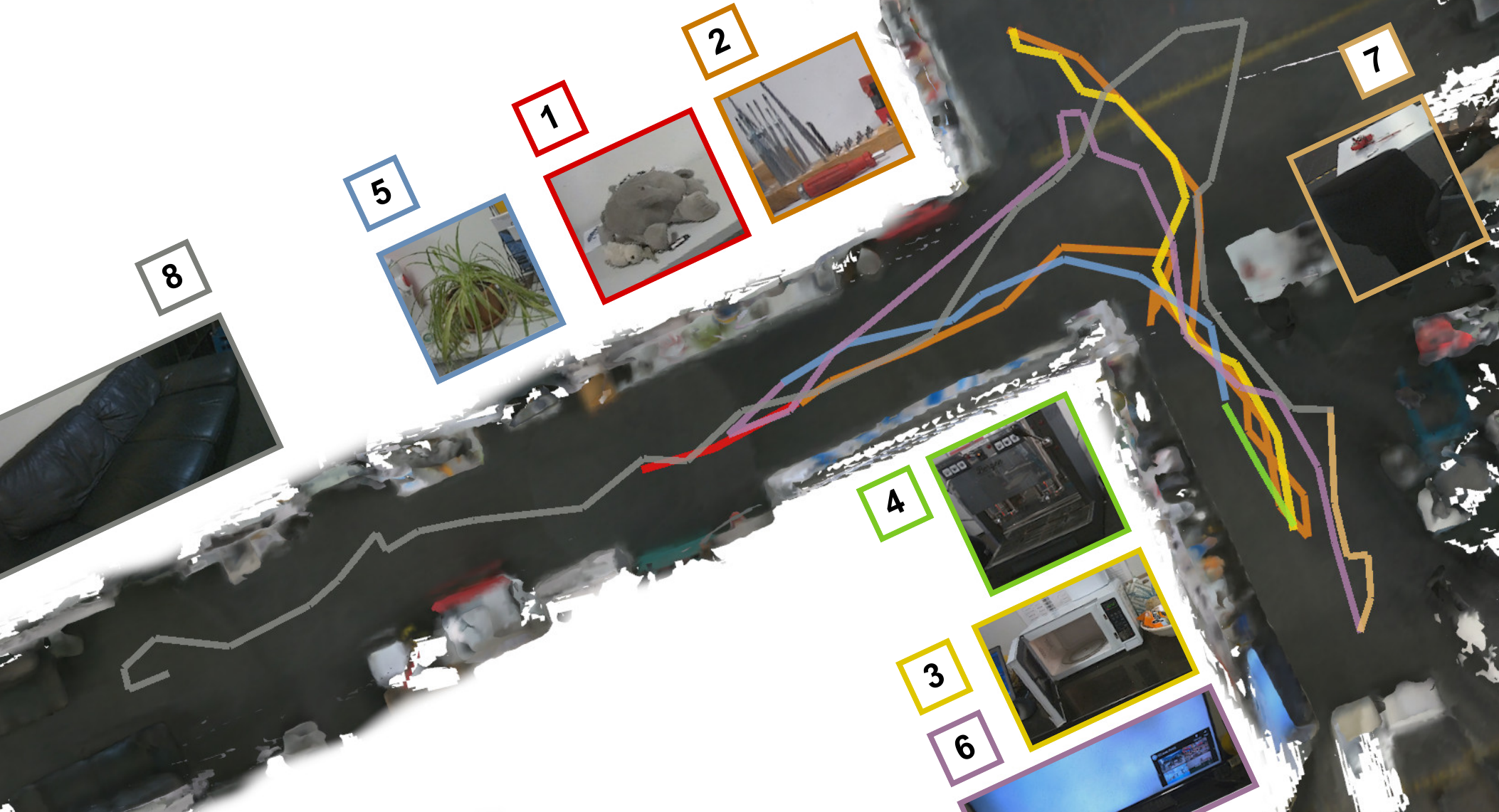}
    \caption{The robot finds eight different objects in succession: a toy elephant, a screwdriver, a microwave oven, a coffee machine, a plant, a TV, a chair, and a sofa. Notice that the eight trajectories are connected. This is because we send the next target immediately when the robot locates the current OOI and the robot can reuse the probability map and ESDF. The OOI in the final RGB image is shown (cropped) and outlined in a color corresponding to its trajectory.}
    \label{fig:all-in-one}
    \vspace{-1.5em}
\end{figure}
\begin{figure*}[t]
    \centering
    \includegraphics[width=0.85\linewidth]{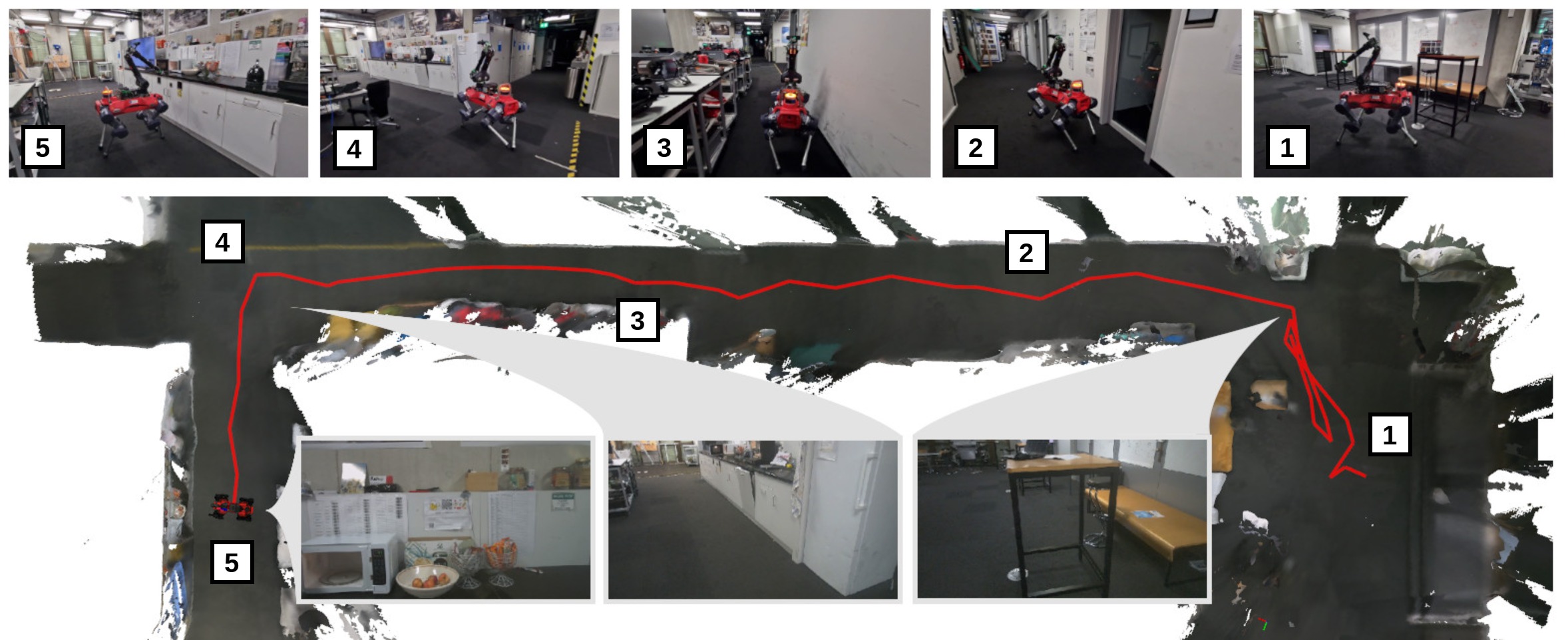}
    \caption{The robot navigates a challenging environment to locate a microwave oven, requiring it to traverse at least 30 meters and make a minimum of two turns. We visualize the camera images in gray rectangular callout.}
    \label{fig:find-microwave-long-distance}
    \vspace{-1.5em}
\end{figure*}

We tested how the common object map can be reused to guide the robot in finding a new OOI efficiently. The robot is given a sequence of tasks, including finding all four basic categories and four open-vocabulary categories in succession. The trajectories and the OOI in the final image are visualized in Fig.~\ref{fig:all-in-one}. The robot successfully found all of them, indicating the robustness of our method. Moreover, the paths to the ``microwave oven,'' ``coffee machine,'' ``plant,'' and ``chair'' are almost optimal because the common object map provides strong guidance. The robot can also traverse a long distance to find a microwave oven, as shown in Fig.~\ref{fig:find-microwave-long-distance}. Our planner also works outdoors, where we tested the robot's capability to find a plant, a ping pong table, or ping pong paddles. 

\subsection{Ablation Study}

To demonstrate the effectiveness of various components, we conducted an ablation study on different modules in mapping, semantic guidance, and planning. The results, indicating performance in the absence of each module, are presented in Table~\ref{tab:ablations-object}.

\begin{table}[h]
  \centering
  \vspace{-0.5em}
  \caption{Ablation study on different modules. }
    \begin{tabular}{l*{5}{c}}
    \toprule
    Ablated Module & SPL $\uparrow$ & Soft SPL  $\uparrow$ & Success  $\uparrow$ \\
    \midrule
    {w/o} Bayes Filter  & 0.28 & 0.41 & 0.45  \\
    \rowcolor{Gray}
    {w/o} Guidance     & 0.31 & 0.43 & 0.50  \\
    {w/o} Local SBP    & 0.28 & 0.41 & 0.47 \\
    \rowcolor{Gray}
    {w/o} Clearance (soft)    & 0.30 & 0.42 & 0.49 \\
    {w/o} Initial Yaw  & 0.30 & 0.40 & 0.53 \\
    \rowcolor{Gray}
    {w/o} Travel Pitch & 0.30 & 0.42 & 0.48 \\  
    \midrule
    IPPON (full)      &  0.34 & 0.46 & 0.54 \\
    \bottomrule
    \end{tabular}
  \label{tab:ablations-object}
  \vspace{-0.5em}
\end{table}

We compared the Bayes filter against naive mapping, which always registers the highest object probability for each category, $p(v \in \mathcal{O} \mid \mathbf{I}_{1:k}) = \max\nolimits_{t = 1, 2, \ldots, k} p(v \in \mathcal{O} \mid \mathbf{I}_{t})$. We observed that the absence of the Bayes filter leads to frequent incorrect terminations. Additionally, SAN~\cite{xu2023side} sometimes fails to segment the OOI with high probability, thus failing to meet the termination criteria. The Bayes filter can effectively mitigate these issues.

With the semantic guidance turned off, all imagined probability defaults to the average probability, $p_{img}(n) = p_{average}$. This results in a 9\% reduction in SPL. Without semantic guidance, we noticed that when the robot sees a mirror and sink in a bathroom, it still frequently reaches inside for a closer inspection, even if the OOI is a bed.


Evaluating the planner without the local SBP, allowing only straight-line connections between nodes, shows the effectiveness of local SBP in informative path planning settings. Further tests demonstrated that penalizing clearance as a soft constraint (instead of hard constraints) in local SBP significantly increases SPL, though at the cost of increasing hits by 61\% (from 1.13 to 1.52 hits per episode). We also discovered that yawing at the beginning or maximizing the pitch movement both contribute to the planner's success.


\section{Conclusion and Discussion}
\label{sec:conclusion}



In this paper, we introduce IPPON, a method that extends the application of informative path planning from exploration and 3D reconstruction to open-vocabulary object goal navigation. By integrating common-sense reasoning from LLMs, the robot achieves a more effective understanding of exploration areas and knows when to abandon the current search zone. Additionally, through the use of 3D object probability mapping, the planner consistently positions the target object within the field of view upon termination. Our evaluation in the Habitat ObjectNav 2023 Challenge demonstrates that our planner significantly outperforms other state-of-the-art methods by a large margin, even as a zero-shot approach. We have extensively validated its performance with real robots across various objects in both indoor and outdoor environments.

However, IPPON currently faces several limitations. First, even though the objects of interest can be open-vocabulary, the list of common objects still needs to be predefined. This may lead to misclassification and incorrect termination signals if an object does not match any provided labels. For common objects, our mapping is performed solely at the object level, without differentiating between a sink in the kitchen and one in the bathroom. Therefore, when the OOI is a toilet, the robot might still attempt to explore the kitchen due to the typical proximity of toilets to sinks. Additionally, the system struggles to detect very small objects, such as a student card.

To address these limitations, we plan to integrate more advanced visual language models, such as GPT-4V, to dynamically expand the list of common objects during exploration. Applying a Bayes filter to dynamically changing categories will also be a key area of investigation. By introducing room-level detection using scene graphs~\cite{gu2024conceptgraphs, hughes2022hydra, wu2021scenegraphfusion} or text-based maps~\cite{zhangtag}, we expect to improve performance as we can integrate fine-grained relationships between rooms and objects. The robotic arm will be used more actively to examine objects at higher elevations (such as a book on a shelf) or to uncover occluded items (like shoes under a table), thus expanding our navigation capabilities into more challenging 3D scenarios.


\bibliographystyle{IEEEtran}  
\bibliography{example}  

\end{document}